\documentclass[11pt]{article}

\usepackage[]{emnlp2021}

\usepackage{times}
\usepackage{latexsym}

\usepackage[T1]{fontenc}

\usepackage[utf8]{inputenc}

\usepackage{microtype}

\usepackage{multirow}
\usepackage{booktabs}
\usepackage{amsmath}
\usepackage{amssymb}
\usepackage{graphicx}

\usepackage{booktabs, multirow, adjustbox}
\usepackage{tablefootnote}
\usepackage{xspace}
\usepackage{makecell}

\usepackage{caption}
\usepackage{subcaption}

\usepackage{amsmath}

\usepackage{nimbusmononarrow}
\usepackage[]{microtype}

\title{ OCR Synthetic Benchmark Dataset for \\ Indic Languages}

\author{
        Naresh Saini$^{{1,2}*}$  \hspace{0.2cm} Promodh Pinto$^{{1,2}*}$ \hspace{0.2cm} Aravinth Bheemaraj$^{1,2}$ \\
        \textbf{Kumar Deepak}$^{1,2}$ \hspace{0.2cm} \textbf{Dhiraj Daga}$^{1,2}$ \hspace{0.2cm} \textbf{Saurabh Yadav}$^{1,2}$ \hspace{0.2cm} \textbf{Srihari Nagaraj}$^{1,2}$
    \\
    $^1$EkStep Foundation,
    $^2$Tarento Technologies
}

\date{}
\begin{document}
\maketitle
\begin{abstract}
We present the largest publicly available synthetic OCR benchmark dataset for Indic languages. The collection contains a total of 90k  images and their ground truth for 23 Indic languages. OCR model validation in Indic languages require a good amount of diverse data to be processed in order to create a robust and reliable model. Generating such a huge amount of data would be difficult otherwise but with synthetic data, it becomes far easier. It can be of great importance to fields like Computer Vision or Image Processing where once an initial synthetic data is developed, model creation becomes easier.
Generating synthetic data comes with the flexibility to adjust its nature and environment as and when required in order to improve the performance of the model. Accuracy for labeled real-time data is sometimes quite expensive while accuracy for synthetic data can be easily achieved with a good score.

\end{abstract}

\section{Introduction}

Synthetic data generation is the process of generating images and corresponding ground truths with different font family, font colour, font size, background color, space between words and many more. Through synthetic data generation we can generate unlimited training data for OCR without any annotators. The major goal of generating synthetic data set is to  validate different OCR models and it can also be used to improve OCR accuracy for different vernacular languages. Pre-trained open source OCR models like tesseract and other deep learning based models are not very accurate on vernacular languages and also not very accurate on scanned and colourful documents irrespective of languages. To increase the accuracy one can fine-tune pre-existing models and for that we need more training data with different variations. We can generate custom data set to generate custom models with higher accuracy. Synthetic data originated in the ’90s, but the real usage occurred in the recent years with people understanding the risks in data science and how fairly that can be eliminated with the usage of synthetic data.

We approach the synthetic image generation problem on an exhaustive coverage of multilingual set of document images. Our work could be used in evaluation of the impact of synthetic data attributes on model performance. We fine tuned the tesseract model for Indian vernacular languages using synthetic data and real data so we were able to achieve around 1 percentage higher accuracy.

OCR (Optical character recognition) performance has increased over the time with the rise of different neural networks. Many of these advances have occurred in constrained settings, where the images are drawn from a single domain and language, such as book scans or scene text. Performance of these models goes down in case of documents of different domains and languages. Performance of the model also deteriorates in case of complex background, new fonts, different lighting conditions and sparse text. If these data variations are not in training data in that case model performance will not be that accurate and it's also difficult to collect and annotate these types of data.  
One of the primary challenges with creating better models for the unconstrained scenario is the lack of annotated training data. Neural models require tens of thousands of annotated images and ground truth text that are both expensive and time consuming to produce.

To solve this problem one of the effective approaches is synthetic data generation.
We can generate lots of data from a few texts and fonts but however, this common approach to synthetic models breaks down when the attributes of the synthetic images do not match the test images. To avoid this issue we generated images which were close to test data.
In this work we have generated synthetic data for 22 Indian languages recognised by the constitution of India. For each language we have generated around 3500 images corresponding to different available fonts in that language. These fonts are both old(Unicode and Non-Unicode) and new(Unicode and Non-Unicode) \cite{DBLP:conf/icdar/EtterRCS19}.

\begin{table*}[htbp]
\caption{Language Wise Font Distribution}
\fontsize{3}{3.5}\selectfont
\resizebox{\textwidth}{!}{%
\begin{tabular}{llll}
\hline 
\textbf{Language}  & \textbf{Fonts}                                                                                                                                                                                                                                                                                                                                                                                                                                                                                                                  & \multicolumn{2}{l}{\textbf{\hspace{0.5cm}{Total Count}}} \\
\textbf{}          & \textbf{}                                                                                                                                                                                                                                                                                                                                                                                                                                                                                                                         & \textbf{Unicode}  & \textbf{Non-Unicode} \\
\hline
\textbf{Hindi}     & \begin{tabular}[c]{@{}l@{}}timesroman, krutidev,helvatica,gargi,aparajita,cambay-italic,akshar non-unicode,samanata,cdacotygn-old-unicode,sakalbharati normal,mukta new \\ non-unicode,kokila,aparajita-old-unicode,sarai,nirmala,cambay-regular,hind-regular,mangal old non-unicode,notosans-regular,shree dv0726 ot -\\ old non-unicode,kalimati,nakula,sarai-07,liberationserif-regular\end{tabular}                                                                                                                           & 20                & 5                    \\\hline
\textbf{Sanskrit}  & \begin{tabular}[c]{@{}l@{}}akshar,cambay-italic,nirmala,nakula,sarai,sarai-07,aparajita-old-unicode,mangal old non-unicode,kokila,lohit devanagari non-unicode,gargi,mukta\\ -regular,notosans-regular,poppins-black,rajdhani-bold,mangal,cambay-regular,samanata,sakalbharati normal,aparajita-old,hind-regular,shree-\\ dv0726 ot old non-unicode,cdacotygn-old,samyak devanagari,kalimati\end{tabular}                                                                                                                         & 21                & 5                    \\\hline
\textbf{Tamil}     & \begin{tabular}[c]{@{}l@{}}liberationserif,helvetica old,symbol old,tscu-saiindira-old,courier old,akshar,tmotabbi-ship-old,muktamalar-bold,kavivanar-regular,baloothambi2\\ -semibold,arial-unicode-ms,pavanam-regular,notoseriftamil-italic-variablefont,arimamadurai-black,latha,catamaran-variablefont-wght,nirmala,lohit-\\ tamil,meerainimai-regular,hindmadurai-semibold,latha old ,tmotabb-ship-old,karlatamilinclined regular\end{tabular}                                                                               & 17                & 3                    \\\hline
\textbf{Telugu}    & \begin{tabular}[c]{@{}l@{}}gautami,helvetica,akshar-old,timesroaman old,courier old,vemana ansi telugu font nonunicode old,notosanstelugu variablefont wdthwght nonunicode \\ hindguntur-bold,gurajada--old,suranna-regular,hindguntur-regular,raviprakash nonunicode old,notoseriftelugu variablefont wght nonunicode,peddana-\\ regular,mandali regular--old,gidugu--old,othana2000 regular nonunicode old,ntr--old,ponnala regular ansi  nonunicode old,ramaraja-regular--old,lohit\\ telugu nonunicode old\end{tabular}       & 10                & 7                    \\\hline
\textbf{Malayalam} & \begin{tabular}[c]{@{}l@{}}liberationserif,helvetica old,timesroman old,meera-regular-old,ml nila04 nonunicode old,suruma nonunicode,rachana-old,kartika old nonunicode,\\ akshar-regular-old-unicode,manjari-regular,lohit-malayalam old nonunicode,baloochettan2-regular,chilanka-regular,anjalioldlipi--old,noto serif\\ malayalam nonunicode\end{tabular}                                                                                                                                                                     & 7                 & 6                    \\\hline
\textbf{Kannada}   & \begin{tabular}[c]{@{}l@{}}helvetica old,malige b nonunicode,arialmt old,kedage n nonunicode,kar kuvempu,lohit kannada nonunicode,malige i nonunicode,tunga-regular,\\ kar k s narasimhaswamy,kar da raa bendre,malige t ,kar puchamthe,kedage i nonunicode,kedage t nonunicode,kar chandrashekhara kambara\\ akshar-regular,kar puthina,kar u r ananthamurthy,kar vi kru gokak,kar shivarama karantha,malige n nonunicode,kar gopalakrishna,adiga,kar\\ girish karnad,kedage b nonunicode\end{tabular}                           & 14                & 9                    \\\hline
\textbf{Oriya}     & \begin{tabular}[c]{@{}l@{}}shreeorix,liberationSans old,arial,mstt31,shree old,arialmt new,akrutioribhuban,SHREE,timesroman old,cairofont nonunicode,or1-jagannatha-old,\\ utkal  old nonunicode,samyak old nonunicode,or14-utkal-old,notosansoriya thin nonunicode,kalinga-old,notosansoriya bold nonunicode,lohit-\\ oriya old nonunicode,baloobhaina2-extrabold,or4-nilachala-old,notosansoriya regular nonunicode,baloobhaina2-regular\end{tabular}                                                                           & 6                 & 6                    \\\hline
\textbf{Punjabi}   & \begin{tabular}[c]{@{}l@{}}helvetica,corbelLight old,arial old,mera,gurbaniAkhar old,puntypo7 nonunicode,muktamahee-regular,lohit nonunicode,notoserifgurmukhi regular-\\ nonunicode,notoserifgurmukhi variablefont wght nonunicode,baloopaaji2-regular,langar-regular,gur old letterpress 4 book nonunicode,dwarka 5-\\ medium nonunicode,baloopaaji2-bold,muktamahee-extrabold,raavi\end{tabular}                                                                                                                               & 6                 & 6                    \\\hline
\textbf{Marathi}   & \begin{tabular}[c]{@{}l@{}}shreeDev,devnagari old,arialUnicode old,shree dv0726 ot old,lohit devanagari nonunicode,sakalbharati normal,aparajita-old,akshar unicode,mukta-\\ regular,notosans regular nonunicode,kokila,cambay-italic,samyak devanagari nonunicode,sarai,cdacotygn-old,mangal nonunicode,mangal old-\\ nonunicode,gargi\end{tabular}                                                                                                                                                                              & 9                 & 6                    \\\hline
\textbf{Gujrati}   & \begin{tabular}[c]{@{}l@{}}albanyamt,shruthi,saral,samayak,padmaa,timesroman old,shruti,rekha,padmaabold,gopika nonunicode,saumil,lohit nonunicode,rasa rajkot old-\\ nonunicode,notosansgujarati black nonunicode,baloobhai2-bold,notoserifgujarati variablefont,hindvadodara-semibold,aakar-old,bhavnagar-old,\\ bhuj-old,muktavaani-regular,rekha,ekatra-old,samyak old nonunicode\end{tabular}                                                                                                                                & 10                & 5                    \\\hline
\textbf{Konkani}   & \begin{tabular}[c]{@{}l@{}}nirmala,shreedev,mangal,arial,brhdevanagari,notosans regular nonunicode,hind-regular,akshar,aparajita-old,cdacotygn-old,lohit devanagari non-\\ unicode,cambay-regular,mangal nonunicode,samyak devanagari nonunicode,mangal old nonunicode,shree dv0726 ot old nonunicode\end{tabular}                                                                                                                                                                                                                & 6                 & 6                    \\\hline
\textbf{Nepali}    & \begin{tabular}[c]{@{}l@{}}preeti, APSDVstardust,chankya old,AakritiCK,kavita,arial,notosansdevanagari,mangal,kokila,kalimati regular old nonunicode,poppins-bolditalic,poppins\\ -extrabolditalic,mukta-bold,poppins-thin,mukta-medium,poppins-mediumitalic,poppins-extralight,poppins-medium,mukta-regular,poppins-black,\\ mukta-semibold,poppins-regular,poppins-italic,poppins-thinitalic,poppins-extralightitalic,rajdhani-semibold,poppins-semibolditalic,poppins-semibold\\ ,rajdhani-regular,poppins-bold\end{tabular}   & 20                & 1                    \\\hline
\textbf{Kashmiri}  & \begin{tabular}[c]{@{}l@{}}mehr nastaliq web 3,alvi nastaleeq regular,alqalam naqsh regular,pak nastaleeq old,jameel noori nastaleeq kasheeda,zohar-old,alqalam taj nastaleeq-\\ regular,aslam,naskh-old,alqalam nabeel regular,jameel-noori-nastaleeq\end{tabular}                                                                                                                                                                                                                                                               & 10                & 1                    \\\hline
\textbf{English}   & \begin{tabular}[c]{@{}l@{}}timesnewroman,bookantiqua nonunicode old,arial,tahoma nonunicode old,constantia nonunicode old,times,calibri,trebuchetms old nonunicode\\ fraunces-9pt-semibolditalic,notosans-bold,courier prime nonunicode,nixieone old nonunicode,timesbd0 nonunicode,exo2-semibold old nonunicode\\ stanberry old nonunicode,zai-courierpolski1941 nonunicode old,opensans-variablefont-wdth,wght nonunicode,ariallgt nonunicode\end{tabular}                                                                      & 4                 & 7                    \\\hline
\textbf{Urdu}      & \begin{tabular}[c]{@{}l@{}}mehr nastaliq web 3,alvi nastaleeq regular,alqalam naqsh regular,pak nastaleeq old,jameel noori nastaleeq kasheeda,zohar-old,alqalam taj nastaleeq-\\ regular,aslam,naskh-old,alqalam nabeel regular,jameel-noori-nastaleeq\end{tabular}                                                                                                                                                                                                                                                               & 10                & 1                    \\\hline
\textbf{Dogri}     & \begin{tabular}[c]{@{}l@{}}sakalbharatidv-normal,gist-mai-dvotdhruv-n-ship-old,notosansdevanagari black,gist-nepotdhruv-n-ship-old,gist-snd-dvotdhruv-n-ship-old,notosans-\\ devanagari- medium,inknutantiqua-medium,gist-kok-dvotdhruv-n-ship-old,notosansdevanagari semibold,gist-snd-dvotvinit-n-ship-old,inknutantiqua\\ -semibold,gist-doi-dvotvinit-n-ship-old\end{tabular}                                                                                                                                                 & 9                 & 3                    \\\hline
\textbf{Sindhi}    & notosansdevanagari black,mangal old,notosansdevanagari medium,gist-snd-dvotvinit,notosansdevanagari semibold,lohit-sindhi-old,gist-snd-dvotdhruv                                                                                                                                                                                                                                                                                                                                                                                  & 3                 & 4                    \\\hline
\textbf{Maithali}  & \begin{tabular}[c]{@{}l@{}}sakalbharatidv-normal,notosansdevanagari black,inknutantiqua-semibold ,gist-nepotdhruv-n-ship-old,notosansdevanagari semibold,gist-doi-dvotvinit-n\\ -ship-old,gist-snd-dvotvinit-n-ship-old,inknutantiqua-medium,gist-mai-dvotdhruv-n-ship-old,notosansdevanagari medium,gist-nepotdhruv-n-ship-old\end{tabular}                                                                                                                                                                                      & 9                 & 3                    \\\hline
\textbf{Manipuri}  & sakalbharatidv-normal,gist-mni-bnotarbindo-n-ship-nonunicode,nirmala--old                                                                                                                                                                                                                                                                                                                                                                                                                                                         & 3                 & 0                    \\\hline
\textbf{Bodo}      & \begin{tabular}[c]{@{}l@{}}gist-snd-dvotdhruv-n-ship-old,gist-snd-dvotvinit-n-ship-old,notosansdevanagari medium,inknutantiqua-semibold,gist-mai-dvotdhruv-n-ship-old,\\ inknutantiqua-medium,gist-doi-dvotvinit-n-ship-old,gist-kok-dvotdhruv-n-ship-old,notosansdevanagari semibold,sakalbharatidv-normal,gist-nepotdhruv\\ -n-ship-old,notosansdevanagari black\end{tabular}                                                                                                                                                   & 9                 & 3                    \\\hline
\textbf{Santali}   & gist-mni-bnotarbindo-n-ship-nonunicode,notosansolchiki regular,sakalbharatidv-normal,nirmala-old                                                                                                                                                                                                                                                                                                                                                                                                                                  & 2                 & 2                    \\\hline
\textbf{Assamese}  & \begin{tabular}[c]{@{}l@{}}balooda2-bold-new-unicode,banikanta-new-unicode,balooda2-regular-new-unicode,notosansbengali variablefont,likhannormal-old-unicode,\\ truetypewriterpolyglott mela old,atma-regular-new-unicode,atma-bold-new-unicode,charukola bl3l old,hindsiliguri-bold-new-unicode,mitra-old\\ -unicode,optimamodoki rgabo old,muktinarrow-old-unicode,hindsiliguri-regular-new-unicode,faulmannfont lg7mw old,jamrulnormal-old-unicode,\\ ani-old-unicode,lohit bengali new,akaashnormal-old-unicode\end{tabular} & 13                & 6  \\\hline                 
\end{tabular}%
}
\end{table*}

\section{Importance of Synthetic Data}
The scope of synthetic data comes with its power of generating features to meet specific needs or conditions which otherwise would not be available in real-world data. When there is not enough of data during testing or when privacy is considered your utmost priority, synthetic data comes to the rescue.
\begin{enumerate}
  \item  In the medical and healthcare sector, synthetic data is used for testing certain medical conditions or cases for which real data is not available.
  \item ML-based Uber and Google’s self-driving cars are trained with the use of synthetic data.
  \item In the financial sector, fraud detection and protection are very critical. New fraudulent cases can be examined with the help of synthetic data.
  \item Synthetic data enables data professionals to access the use of centrally recorded data while still maintaining the confidentiality of the data. Synthetic data comes with the power to replicate the important features of real data without exposing the true sense of it, thereby keeping privacy intact.
  \item In the research department, synthetic data helps you develop and deliver innovative products for which necessary data otherwise might not be available.
  \item In OCR synthetic data is used for training and fine tuning different OCR models.
\end{enumerate}

\section{Approach}

\subsection{Synthetic Data Generation}
Our synthetic data generation tool renders images at the line level. It has many abilities such as to limit the number of words, identify the source language to support the specific language and remove if found any other languages and break a long text or document into multiple lines.

A list of lines are given through the text corpus and the collected text corpus may be flawed, which can include other languages when it is supposed to be a specific language. In this case, our synthetic generation tool is capable of detecting the source language  and resume the process after having made sure of the specific language that is needed.

More variance is included in the text, with respect to the number of words, space between the words, support of other languages and special characters.

Text can be long and it will have lots of words or sentences which will make the image less realistic as the goal is to generate sentences.  
List of data required for generation of synthetic data:
\begin{itemize}
\item  Text corpus
\item  Fonts
\item  Background image
\item  Noise filters
\item  Text color codes
\end{itemize}
These are the key files that help generate synthetic data. The power of generating features to meet specific needs and conditions that otherwise would not be possible in real world data and that power comes with the synthetic data generation.
Synthetic data generation helps in testing certain conditions and cases for which real data does not exist.
Synthetic data also can replicate the important features of real world data without exposing the true sense of it.
\textbf{\emph{\large{Text Corpus Sample:}}}
\begin{center}
    \includegraphics[width=0.5\textwidth]{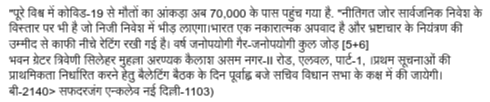}
\end{center}

\subsection{Style Attributes}
Style attributes are the features that are used to render our text into synthetic images. This includes features such as font color, background color, noise features and background images.

The key idea is to get the ground truth with the uniqueness of each text from the text corpus and apply features with different variations that meet with our expectations that help generate synthetic data.
The different types of fonts are collected specific to the language and tested to avoid redundancy.
While processing the text corpus, the text may contain other language texts, in that case the text is being removed.
Different types of background images specific to the needs are collected.

For each language, up to 15 fonts are tested, and the fonts that render the text correctly are chosen for synthetic data generation.
Various types of noise filters are added during the generation to replicate different types of classes in quality.

Use of different text colors is also included to add more variations.
The data is purely synthetic and does not have anything to do with the original data.
For each image, a series of features are generated on a random basis from the collected features that meet with our expectations. Then the series of these features are mapped to the other features and the ground truth in order to generate fully synthetic data.


\begin{table*}[htbp]
\caption{Language Wise Data Distribution}
\fontsize{30}{40.5}\selectfont
\centering
\resizebox{\textwidth}{!}{%
\begin{tabular}{lllllllllllllllll}
\hline\hline 
\textbf{Languages}                  & \textbf{} & \textbf{Class 1} & \textbf{Class 2} & \textbf{Class3} & \textbf{Noise-Added-Real} & \textbf{Scene-Text} & \textbf{} &\vline \textbf{} & \textbf{} & \textbf{Languages}                 & \textbf{} & \textbf{Class 1} & \textbf{Class 2} & \textbf{Class3} & \textbf{Noise-Added-Real} & \textbf{Scene-Text} \\\hline
\multirow{2}{*}{\textbf{English}}   & Synthetic & 365              & 1709             & 1220            & -                         & -                   &          \textbf{} &\vline \textbf{} & \textbf{} & \multirow{2}{*}{\textbf{Gujarati}} & Synthetic & 489              & 2039             & 1504            & -                         & -                   \\
                                    & Real      & 176              & 86               & 87              & 350                       & 38                  &           \textbf{} &\vline \textbf{} & \textbf{} &                                    & Real      & 344              & 79               & 413             & 836                       & 26                  \\\hline
\multirow{2}{*}{\textbf{Hindi}}     & Synthetic & 531              & 2076             & 1525            & -                         & -                   &           \textbf{} &\vline \textbf{} & \textbf{} & \multirow{2}{*}{\textbf{Sanskrit}} & Synthetic & 317              & 1272             & 990             & -                         & -                   \\
                                    & Real      & 488              & 127              & 301             & 916                       & 42                  &           \textbf{} &\vline \textbf{} & \textbf{} &                                    & Real      & 0                & 0                & 0               & 0                         & 0                   \\\hline
\multirow{2}{*}{\textbf{Kannada}}   & Synthetic & 258              & 1460             & 1296            & -                         & -                   &           \textbf{} &\vline \textbf{} & \textbf{} & \multirow{2}{*}{\textbf{Odiya}}    & Synthetic & 240              & 1072             & 778             & -                         & -                   \\
                                    & Real      & 206              & 342              & 0               & 549                       & 28                  &           \textbf{} &\vline \textbf{} & \textbf{} &                                    & Real      & 448              & 112              & 5               & 0                         & 0                   \\\hline
\multirow{2}{*}{\textbf{Tamil}}     & Synthetic & 358              & 1425             & 1328            & -                         & -                   &           \textbf{} &\vline \textbf{} & \textbf{} & \multirow{2}{*}{\textbf{Marathi}}  & Synthetic & 340              & 1507             & 1493            & -                         & -                   \\
                                    & Real      & 240              & 796              & 181             & 1217                      & 40                  &           \textbf{} &\vline \textbf{} & \textbf{} &                                    & Real      & 478              & 255              & 0               & 734                       & 0                   \\\hline
\multirow{2}{*}{\textbf{Malayalam}} & Synthetic & 376              & 1335             & 1181            & -                         & -                   &           \textbf{} &\vline \textbf{} & \textbf{} & \multirow{2}{*}{\textbf{Konkani}}  & Synthetic & 314              & 1370             & 1038            & -                         & -                   \\
                                    & Real      & 369              & 90               & 8               & 468                       & 31                  &           \textbf{} &\vline \textbf{} & \textbf{} &                                    & Real      & 233              & 33               & 8               & 274                       & 0                   \\\hline
\multirow{2}{*}{\textbf{Telugu}}    & Synthetic & 419              & 1888             & 1519            & -                         & -                   &           \textbf{} &\vline \textbf{} & \textbf{} & \multirow{2}{*}{\textbf{Sindhi}}   & Synthetic & 379              & 1651             & 1272            & -                         & -                   \\
                                    & Real      & 829              & 807              & 92              & 0                         & 44                  &          \textbf{} &\vline \textbf{} & \textbf{} &                                    & Real      & 0                & 0                & 0               & 0                         & 0                   \\\hline
\multirow{2}{*}{\textbf{Bengali}}   & Synthetic & 475              & 1907             & 1421            & -                         & -                   &           \textbf{} &\vline \textbf{} & \textbf{} & \multirow{2}{*}{\textbf{Maithali}} & Synthetic & 335              & 1453             & 1117            & -                         & -                   \\
                                    & Real      & 224              & 549              & 81              & 855                       & 32                  &           \textbf{} &\vline \textbf{} & \textbf{} &                                    & Real      & 0                & 0                & 0               & 0                         & 0                   \\\hline
\multirow{2}{*}{\textbf{Santali}}   & Synthetic & 380              & 1708             & 1268            & -                         & -                   &           \textbf{} &\vline \textbf{} & \textbf{} & \multirow{2}{*}{\textbf{Assamese}} & Synthetic & 344              & 1003             & 1106            & -                         & -                   \\
                                    & Real      & 0                & 0                & 0               & 0                         & 0                   &           \textbf{} &\vline \textbf{} & \textbf{} &                                    & Real      & 0                & 0                & 0               & 0                         & 0                   \\\hline
\multirow{2}{*}{\textbf{Urdu}}      & Synthetic & 417              & 1819             & 1459            & -                         & -                   &           \textbf{} &\vline \textbf{} & \textbf{} & \multirow{2}{*}{\textbf{Punjabi}}  & Synthetic & 444              & 1914             & 1460            & -                         & -                   \\
                                    & Real      & 0                & 0                & 0               & 0                         & 0                   &           \textbf{} &\vline \textbf{} & \textbf{} &                                   & Real      & 152              & 60               & 12              & 225                       & 0                   \\\hline
\multirow{2}{*}{\textbf{Nepali}}    & Synthetic & 298              & 1376             & 1029            & -                         & -                   &           \textbf{} &\vline \textbf{} & \textbf{} & \multirow{2}{*}{\textbf{Manipuri}} & Synthetic & 424              & 1987             & 1398            & -                         & -                   \\
                                    & Real      & 430              & 64               & 63              & 558                       & 0                   &           \textbf{} &\vline \textbf{} & \textbf{} &                                    & Real      & 0                & 0                & 0               & 0                         & 0                   \\\hline
\multirow{2}{*}{\textbf{Kashmiri}}  & Synthetic & 264              & 1268             & 937             & -                         & -                   &           \textbf{} &\vline \textbf{} & \textbf{} & \multirow{2}{*}{\textbf{Bodo}}     & Synthetic & 180              & 744              & 531             & -                         & -                   \\
                                    & Real      & 0                & 0                & 0               & 0                         & 0                   &          \textbf{} &\vline \textbf{} & \textbf{} &                                    & Real      & 0                & 0                & 0               & 0                         & 0                   \\\hline
\multirow{2}{*}{\textbf{Dogri}}     & Synthetic & 493              & 2301             & 1660            & -                         & -                   &           \textbf{} &\vline \textbf{} & \textbf{} &                                    &           &                  &                  &                 &                           &                     \\
                                    & Real      & 0                & 0                & 0               & 0                         & 0                   &           \textbf{} &\vline \textbf{} & \textbf{} &                                    &           &                  &                  &                 &                           &                    
\end{tabular}%
}
\end{table*}

\subsection{Class Attributes}
Synthetic data is key in having the data diversity to represent the real world and reduce the bias to solve issues.
Because the synthetic data is automatically labeled and can deliberately include rare but crucial cases.

Types of classes:
\begin{itemize}
\item \hspace{0.06cm}    Class 1
\item  \hspace{0.06cm} Class 2
\item \hspace{0.06cm} Class 3
\end{itemize}

Class is the data quality measure of the condition based on factors such as accuracy, completeness, consistency, reliability and which represents the types of various real world data.

As a first step towards determining the data quality levels, for each data, the series of features that are generated on a random basis in order to map with the other features and ground truth, these features and the mapping information helps determine the quality of the data.

{\large{Class 1:}}

Data accuracy is a key attribute of class 1 type data of high quality.

This is the most basic OCR complexity, with clear text on a white background and font variation.

The aspects such as completeness, consistency, clear visibility and readability are considered in this type of classification.

The methodology in achieving this is by determining the features that are generated on a random basis.

The feature mapping in which the features are aligning with respect to most of the types that are included:
\begin{itemize}
\item   High quality
\item   White background
\item   No noise filter
\item   No color text
\item   Clear background 
\end{itemize}
{\emph{\large{\textbf{Sample Images: }}}}
\begin{center}
    \includegraphics[width=.5\textwidth]{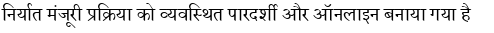}
    \includegraphics[width=.5\textwidth]{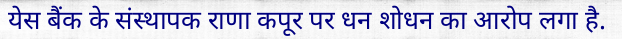}
    \includegraphics[width=.5\textwidth]{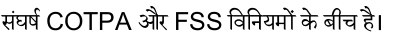}
\end{center}

{\large{Class 2:}}

The goal is to simulate high-quality scanned documents with variations in text color, background color, and font.
In classification of the type class 2, accuracy may not be the key but still however, it should be easily readable with very less of a noise added along with the color text and some less complex background images.
The features that are mostly mapped to generate the type class 2 are listed below:
\begin{itemize}
\item  Average quality
\item  color or White background
\item  Noise filters with minimal changes
\item  color text
\item  Less noise backgrounds  
\end{itemize}
{\emph{\large{\textbf{Sample Images: }}}}
\begin{center}
    \includegraphics[width=.5\textwidth]{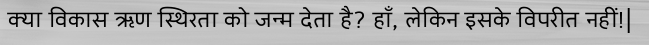}
    \includegraphics[width=.5\textwidth]{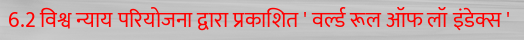}
    \includegraphics[width=.5\textwidth]{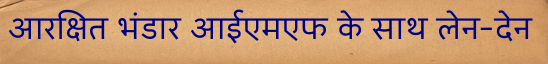}
\end{center}

{\large{Class 3:}}

The goal is to simulate poor-quality scanned documents, complex backgrounds, font variation, noise, and blurring.

In classification of the type class 3, in simplest terms,  poor data quality of inaccurate information.

It lacks accuracy, completeness, visibility, readability and quality.

The feature mapping in which the features are aligning with respect to most of the types that are included:
\begin{itemize}
\item  Bad quality
\item  White or color backgrounds
\item  Heavy noise filters
\item  color text
\item  Complex backgrounds 
\end{itemize}
{\emph{\large{\textbf{Sample Images: }}}}
\begin{center}
    
    \includegraphics[width=.5\textwidth]{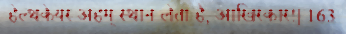}
    \includegraphics[width=.5\textwidth]{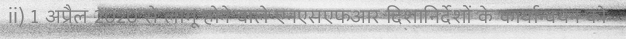}
    \includegraphics[width=.5\textwidth]{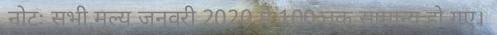}
\end{center}

\subsection{Font Attributes}
It is a key challenge when it comes to identifying the fonts for each specific language. We were able to gather various fonts specific to language.

One of the key challenges include in identifying the fonts as well as validating the gathered fonts.

The gathered fonts are separated into new and old fonts, also as Unicode and Non-Unicode fonts.

During validation, issues to look for are, whether it supports alphanumeric and special characters and multi language support as well. 
{\emph{\large{\textbf{Unicode Fonts: }}}}
\begin{center}
    \includegraphics[width=.5\textwidth]{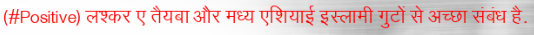}
    \includegraphics[width=.5\textwidth]{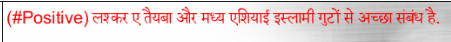}
\end{center}
{\emph{\large{\textbf{Non-Unicode Fonts: }}}}
\begin{center}
    \includegraphics[width=.5\textwidth]{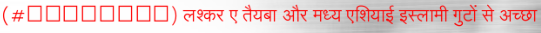}
    \includegraphics[width=.5\textwidth]{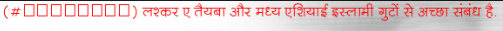}
\end{center} 
\subsection{Noise Filter Attributes}
The noise filters help represent the synthetically generated data to not have uniform variance with respect to quality and pixels.

The real data of different quality such as good, average and bad. Our synthetic data generation tool has the ability to add such noise to represent the data in different quality.

{\emph{\large{\textbf{Sample Images: }}}}
\begin{center}
    \includegraphics[width=.5\textwidth]{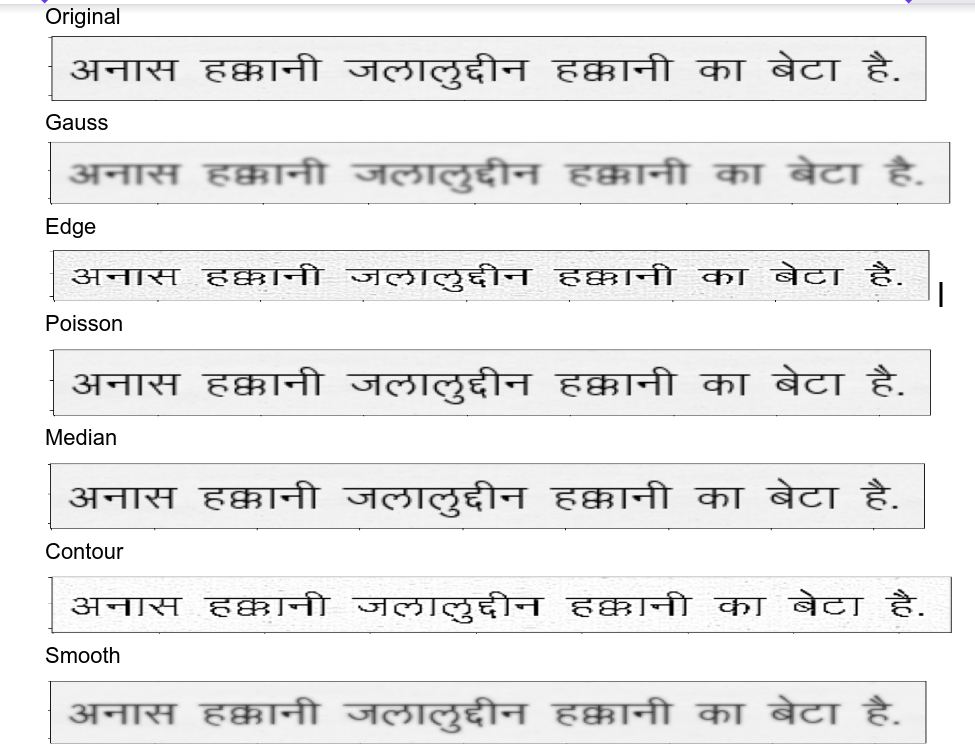}
\end{center}

\section{Scene-Text}
Scene-Text images are taken from the natural scenes which are captured by the camera. Scene-Text images includes complex background and different text appearances.
The text in natural scene images can be in different languages,colour and fonts. 
\\
Scene-Text data is available in 8 languages that are English, Hindi, Tamil, Telugu, Malayalam, Kannada, Bengali and Gujarati.\\
For a given image or a scene, the lines  are localized by bounding boxes and that region is cropped to an image.
Using OCR the text present in the collected images are extracted to a machine-readable format that we can edit, search, and can use for further processing.\\
Most of the OCR models does not perform well on scene text data so benchmark of the OCR model on scene text data would be a good parameter.
{\emph{\large{\textbf{Sample Images: }}}}
\begin{center}
    \includegraphics[width=.5\textwidth]{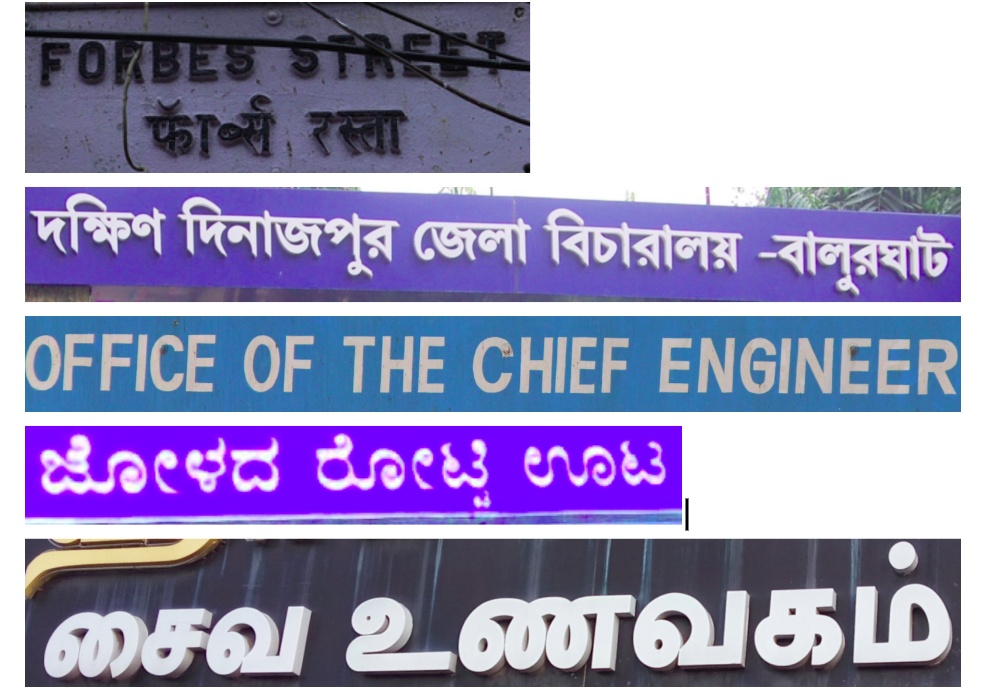}
\end{center}

\section{Coverage of Characters Sets}
The objective is to achieve the coverage of all characters and numbers while or after generating the dataset.
From the generated dataset, the missing characters and the numbers were identified through analysis at font level. The data has been generated across various fonts and under those each font the existence of characters and numbers were identified through analysis and those specific characters and numbers set were generated synthetically and added back to the dataset in order to achieve character coverage for every font that is included in the dataset for all the existing languages.

\section{Dataset}
To generate synthetic images, our tool uses seed text sentences drawn from various domains such as legal, education, govt orders and old documents ( before the 1960s ). Some part of the data was downloaded from the Anuvaad parallel corpus \cite{DBLP:journals/corr/abs-2104-05596} and also we were able to generate it from the google vision api. 

Google vision API was used for OCR. As we know google vision is also not completely accurate so we manually corrected and extracted text \cite{DBLP:conf/icdar/ShenZDLCL21}. We collected documents from various domains like legal, NCERT books, govt orders and old documents. 

Text corpus can also be filtered further on the basis of language specific characters like special character, digits, auxiliary and any other characters.
Some of those lines that contain those certain characters, with the help of regular expressions, can be drawn.

For each image, a series of features are generated on a random basis from the collected features that meet with our expectations.

For each and every character set, we can define percentage value so that we can filter text corpus in only the required ratio.


Good to considered that it would be better to include a small portion of the real data as well to bring in the balance by adding some of the hardest features that are otherwise challenging to generate synthetically.
By adding a small portion of the real data along with the synthetic data, the real data represents some of the key characteristics of the real world use, that would also help the data training achieve better results.

\begin{table*}[htbp]
\fontsize{10}{12.5}\selectfont
\caption{Language Wise Script Details}
\vspace{0.3cm}

\resizebox{\textwidth}{!}{%
\begin{tabular}{llllllll}
\hline\hline 
\textbf{Language} & \textbf{Script}                   & \textbf{} & \textbf{} &\vline \textbf{} & \textbf{} & \textbf{Language} & \textbf{Script}
\\ \hline
Bengali           & Bangla, Devanagari, Indic, Brahmi &           &           & \vline          &           & Marathi           & Devanagari, Indic    \\
Assamese          & Bangla, Devanagari, Indic, Brahmi &           &           &\vline           &           & Urdu              & Arabic               \\
Bodo              & Devanagari, Indic                 &           &           &\vline           &           & Kashmiri          & Perso Arabic         \\
Dogri             & Devanagari, Indic, Dogra          &           &           &\vline           &           & Kannada           & Kannada, Indic       \\
English           & Latin                             &           &           & \vline          &           & Santali           & Ol Chiki             \\
Gujarati          & Devanagari, Indic, Gujarati       &           &           & \vline          &           & Sanskrit          & Devanagari, Indic    \\
Konkani           & Devanagari, Indic                 &           &           & \vline          &           & Nepali            & Devanagari           \\
Tamil             & Tamil, Brahmi, Indic              &           &           &  \vline         &           & Oriya             & Kalinga, Odia, Indic \\
Malayalam         & Malayalam, Indic                  &           &           &  \vline         &           & Punjabi           & Gurmukhi, Indic      \\
Telugu            & Telugu, Brahmi, Indic             &           &           &   \vline        &           & Sindhi            & Devanagari           \\
Hindi             & Devanagari, Indic                 &           &           &\vline           &           & Manipuri          & Devanagari           \\
Maithili          & Devanagari                        &           &           & \vline          &           &                   &   \\\hline                  
\end{tabular}%
}
\end{table*}

\section{Text Corpus Source and License}
The text corpus used for synthetic data generation is mostly made from the sources
While some are available under a Creative Commons license and some are available under other public open data standard licenses.

\begin{itemize}
\item https://github.com/project-anuvaad/anuvaad-parallel-corpus
\item https://legislative.gov.in/regional-language
\item https://ncert.nic.in/textbook.php
\item https://main.sci.gov.in/vernacular\_judgment
\item https://archive.org/
\end{itemize}

\section{Challenges}
Synthetic data has strong roots in Artificial Intelligence with numerous benefits but still has some challenges which need to be taken care of while dealing with synthetic data. These are as follows:
Difficulty in generating synthetic data.
Dealing with the number of inconsistencies encountered while replicating the complexities from real data to synthetic data.
The flexible nature of synthetic data makes it biased in behavior while validation or training.
Validating it with synthetic test data might not be enough for users and might be required to validate it with real data.
There could be some hidden follies on the performance of algorithms trained with simplified representations of synthetic data which lately may pop out while dealing with real data.
Users in some cases may not accept synthetic data to be a valid data.
Generating all necessary features from real data might become complex in nature and also there can be a possibility of missing out on some necessary features during this generation procedure.

\section{Future Work}
This work describes the first stages of our work in synthetic data generation. In this we presented a dataset with approximately 2500 records of lines for each 23 languages.
The dataset mostly contains synthetically generated images, a small portion of real world data and scene-text data in order to bring in the balance by adding some of the hardest features that are otherwise challenging to  generate synthetically.
Since the real data and scene-text data represents some of the key characteristics of the real world use and as well as will help in maintaining the data diversity, more of these data will be added in future as currently it is not available for some of the low resource languages.
For each image a series of features such as noise filters, text color, simple and complex backgrounds were added to meet with our expectations. More such features will be added to closely represent the real world data.
We will investigate the synthetic image generation problem on a large multilingual set of unconstrained document images and present a comprehensive evaluation of the impact of synthetic data attributes on model performance.

Finally we intend to extend our dataset with more vernacular languages and handwritten synthetic data generation.

\section{Conclusion}
{We presented the approach of synthetic data generation and impact of synthetic data generation on OCR model performance and their benchmarking. In this approach we generated image data for the challenging Indian languages.The results show that with a relatively small number of fonts and text, a model can be trained or fine-tuned that is competitive or outperforms a model trained on real images.
}
\section*{Acknowledgments}
All the authors acknowledge the support of \textbf{EkStep Foundation} and \textbf{Tarento Technologies} in presenting this work. We would like to thank Sumanth and Gowtham from IIT Madras, who have helped us shape this paper. We would also like to thank Vivek Raghavan for technical support and guidance.

\bibliography{anthology,custom}
\bibliographystyle{acl_natbib}
\nocite{DBLP:journals/symmetry/SporiciCB20}
\nocite{6240859}
\nocite{mathew_jain_jawahar_2017}
\nocite{bhattacharya_parui_mondal_2009}

\end{document}